# Intelligent bidirectional rapidly-exploring random trees for optimal motion planning in complex cluttered environments*

Ahmed Hussain Qureshi[1,2], Yasar Ayaz[1]

[1]Robotics And Intelligent Systems Engineering (RISE) Lab
Department of Robotics and Artificial Intelligence
School of Mechanical And Manufacturing Engineering (SMME)
National University of Sciences And Technology (NUST)
H-12 Campus, Islamabad, 44000, Pakistan.

[2]Department of System Innovation
Graduate School of Engineering Science
Osaka University, 1-3 Machikaneyama, Toyonaka, Osaka, Japan.

March 27, 2017

## Abstract

The sampling based motion planning algorithm known as Rapidly-exploring Random Trees (RRT) has gained the attention of many researchers due to their computational efficiency and effectiveness. Recently, a variant of RRT called RRT* has been proposed that ensures asymptotic optimality. Subsequently its bidirectional version has also been introduced in the literature known as Bidirectional-RRT* (B-RRT*). We introduce a new variant called Intelligent Bidirectional-RRT* (IB-RRT*) which is an improved variant of the optimal RRT* and bidirectional version of RRT* (B-RRT*) algorithms and is specially designed for complex cluttered environments. IB-RRT* utilizes the bidirectional trees approach and introduces intelligent sample insertion heuristic for fast convergence to the optimal path solution using uniform sampling heuristics. The proposed algorithm is evaluated theoretically and experimental results are presented that compares IB-RRT* with RRT* and B-RRT*. Moreover, experimental results demonstrate the superior efficiency of IB-RRT* in comparison with RRT* and B-RRT in complex cluttered environments.

## 1 Introduction

Motion planning is a well-known problem in robotics [24]. It can be defined as the process of finding a collision-free path for a robot from its initial to goal point while avoiding collisions with any static obstacles or other agents present in its environment. Although motion planning is not the only fundamental problem of robotics, perhaps it has gained popularity among researchers due to widespread applications such as in robotics [21], assembly maintenance [3], computer animation [7], computer-aided surgery [9], manufacturing [22], and many other aspects of daily life.

The journey of finding solution to motion planning problems started with complete planning algorithms

*This is the authors' version of the paper published in Elsevier Robotics and Autonomous Systems Journal. The source code of this paper is available at: github.com/ahq1993.

that comprised of deterministic path planning approach. Complete motion planning algorithms [35] [28] are those algorithms that converges to a path solution, if one exists, in finite time. These algorithms are proven to be computationally inefficient [2] in most of the practical motion planning problems [13]. Resolution complete algorithms were then introduced that require fine tuning of resolution parameters for providing the motion planning solution, if one exists, in a finite time period. Artificial Potential Fields (APF) [16] is a well-known resolution complete algorithm. However, APF suffers from the problem of local minima [18] and does not performed well in the environment with narrow passages. Hence, the search for an efficient solution to the problem continued and the idea of exact roadmaps was introduced in the literature which relies on the discretization of the given search space. This discretization of search space makes the algorithm computationally expensive for higher dimensional spaces, that is why the application of such algorithms like Cell Decomposition methods [19] [1], Delaunay Triangulations [10] and Dynamic Graph Search methods [4] [25] are limited to low dimensional spaces only [5]. Moreover the algorithms that combine the set of allowed motions with the graph search methods thus generating state lattices, such as in [8] [31] [30], also suffered from the undesirable effects of discretization. Hence to solve the higher dimensional planning problems, the sampling-based algorithms were introduced [5]; the main advantage of sampling-based algorithms as compared to other state-of-the-art algorithms is avoidance of explicit construction of obstacle configuration space. These algorithms ensure *probabilistic completeness* which implies that as the number of iterations increases to infinity, the probability of finding a solution, if one exists, approaches one. The sampling-based algorithms have proven to be computationally efficient [2] solution to motion planning problems. Arguably, the most well-known sampling-based algorithms include Probabilistic Road Maps (PRM) [14] [15] and Rapidly exploring Random Trees (RRT) [23]. However, PRMs tend to be inefficient when obstacle geometry is not known beforehand [13]. Therefore, in order to derive efficient solutions for motion planning in the practical world, the Rapidly-exploring Random Trees (RRT) algorithms [23] have been extensively explored. Various algorithms enhancing original RRT algorithm have been proposed [26], [6], [20], [13]. These algorithms present a solution regardless of whether specific geometry of the obstacles is known beforehand or not. One of the most remarkable variant of RRT algorithm is RRT*, an algorithm which guarantees eventual convergence to an optimal path solution [13], unlike the original RRT algorithm. Just like the RRT algorithm, RRT* is able to generate an initial path towards the goal very quickly. It then continues to refine this initial path in successive iterations, eventually returning an optimal or near optimal path towards the goal as the number of iterations approach infinity [12]. This additional guarantee of optimality makes the RRT* algorithm very useful for real-time applications [29]. However, some major constraints still exist in this RRT variant which are presented in this paper. For example:

(i) its slow convergence rate in achieving the optimal solution;

(ii) its significantly large memory requirements due to the large number of iterations utilized to calculate the optimal path; and

(iii) its rejection of samples which may not be directly connectable with the existing nodes in the tree, but may lie closer to the goal region and hence could aid the algorithm in determining an optimal path much faster.

Various heuristics have been introduced, such as [32] [33] [34] [17], which perform guided search of the given space instead of pure uniform search (as by RRT and RRT*). Although these biased sampling heuristics make the original RRT* algorithm fast but there is a drawback of computational overload caused by biased sampling. This computational overload limits their application to a limited number of fields [27]. Moreover another disadvantage of deterministic sampling heuristics is that they may interfere with the algorithm characteristics. For example assume a simple case of using goal-biased sampling [33] with bidirectional RRT that alternatively grows two trees. The use of this biased sampling will cause the two trees to always remain in one half of the search space, which is quite undesirable. Hence, to cover the whole search space, a separate sample generator is required for both trees which will cost a significant computational load. Hence there is a need of some better approach that enhances the convergence rate of RRT* for achieving the optimal path solution without affecting the randomization of its sampling heuristic. More recent proposition is the bidirectional version of RRT* known as B-RRT* [11]. B-RRT* presented in [11] is a simple bidirectional implementation of RRT*. B-RRT* uses a slight variation of greedy RRT-Connect heuristic [20] for the connection of two trees. Two directional trees employing greedy connect heuristic for the connection of trees does not ensure *asymptotic optimality* [11]. The B-RRT* uses slight variation of greedy heuristic i.e., the tree under process first searches for the neighbor vertices before making an attempt to connect the trees using RRT-Connect heuristic [20]. This hybrid greedy connection heuristic of B-RRT* slows down its ability to converge to the optimal solution and also makes it computationally expensive. More detailed discussion is provided in the analysis section. This paper introduces a bidirectional variation to the RRT* algorithm, with unique sample insertion and tree connection heuristics that allows fast convergence to the optimal path solution. The proposed Intelligent Bidirectional-RRT* (IB-RRT*) algorithm has been tested for its robustness in both 2-D and 3-D environments and has also been compared with other state-of-art algorithms such as Bidirectional-RRT*[11] and RRT* itself [12]. The rest of the paper is organized as follows. Section 2 addresses the problem definition, Section 3 explains the RRT* algorithm while Section 4 describes the B-RRT* motion planning algorithm in detail. Section 5 presents the proposed Intelligent Bidirectional-RRT* (IB-RRT*). Section 6 presents analysis of the three algorithms under investigation in terms of probabilistic completeness, asymptotic optimality, convergence to the optimal solution and computational complexity. Section 7 provides experimental evidence in support of theoretical results presented in the previous section, whereas Section 8 concludes the paper, also suggesting some future areas of research in this particular domain.

# 2 Problem Definition

Let the given state space be denoted by a set $X \subset \mathbb{R}^n$, where $n$ represents the dimension of the given space i.e., $n \in \mathbb{N}$, $n \geq 2$. The configuration space is further classified into obstacle and obstacle-free regions denoted by $X_{\mathrm{obs}} \subset X$ and $X_{\mathrm{free}} = X \backslash X_{\mathrm{obs}}$, respectively. $X_{\mathrm{goal}} \subset X_{\mathrm{free}}$ is the goal region. Let $T_{\mathrm{a}} = (V_{\mathrm{a}}, E_{\mathrm{a}}) \subset X_{\mathrm{free}}$ and $T_{\mathrm{b}} = (V_{\mathrm{b}}, E_{\mathrm{b}}) \subset X_{\mathrm{free}}$ represent two growing random trees, where $V$ denotes the nodes and $E$ denotes the edges connecting these nodes. $x^{\mathrm{a}}_{\mathrm{init}} \in X_{\mathrm{free}}$ and $x^{\mathrm{b}}_{\mathrm{init}} \in X_{\mathrm{goal}}$ represent the starting states for $T_{\mathrm{a}}$ and $T_{\mathrm{b}}$. The function $\mu()$ computes the Lebesgue measure of any given state space e.g. $\mu(X)$ denotes the Lebesgue measure of the whole state space $X$. It is also called the $n$-dimensional volume of any given configuration. This paper only considers Euclidean space and positive Euclidean distance between any two states e.g., $x_1 \in X$ and $x_2 \in X$ is denoted by $\mathrm{d}(x_1, x_2)$. The closed ball region of radius $r \in \mathbb{R}, r > 0$ centered at $x$ is denoted as $\mathfrak{B}_{x,r} := \{x_2 \in X : d(x, x_2) \leq r\}$, where $x \in X$ can be any given configuration state. Let the path connecting any two states $x_1 \in X_{\mathrm{free}}$ and $x_2 \in X_{\mathrm{free}}$ be denoted by $\sigma : [0, s']$, such that $\sigma(0) = x_1$ and $\sigma(s') = x_2$, whereas, $s'$ is the positive scalar length of the path. The set of all collision-free paths $\sigma$ is denoted as $\sum_{\mathrm{free}}$. Given any random state $x \in X_{\mathrm{free}}$, the path function connecting initial state $x_{\mathrm{init}}$ and random state $x$ is denoted as $\sigma'_{\mathrm{a}}[0, s_{\mathrm{a}}] \subset X_{\mathrm{free}} | \{\sigma'_{\mathrm{a}}(0) = x_{\mathrm{init}}$ and $\sigma'_{\mathrm{a}}(s_{\mathrm{a}}) = x\}$, while the path function connecting random state $x$ and goal region $X_{\mathrm{goal}}$ is denoted as $\sigma'_{\mathrm{b}}[0, s_{\mathrm{b}}] \subset X_{\mathrm{free}} | \{\sigma'_{\mathrm{b}}(0) = x$ and $\sigma'_{\mathrm{b}}(s_{\mathrm{b}}) \in X_{\mathrm{goal}}\}$. The complete, end-to-end path function i.e., the path function from root to the goal is denoted by $\sigma'_{\mathrm{f}}(\mathrm{s}) = \sigma'_{\mathrm{a}} | \sigma'_{\mathrm{b}} : [0, s] \in X$, where $s$ represents the scalar length of the end-to-end path. The expression $\sigma'_{\mathrm{a}} | \sigma'_{\mathrm{b}} \in X$ describes the concatenation of the two path functions, $\sigma'_{\mathrm{a}}$ and $\sigma'_{\mathrm{b}}$. The path function $\sigma_{\mathrm{f}}$ is the end-to-end feasible path in obstacle-free configuration space, i.e., $\sigma_{\mathrm{f}} \in X_{\mathrm{free}}$. The set of all end-to-end collision-free paths is denoted as $\sum_{\mathrm{f}}$ i.e., $\sigma_{\mathrm{f}} \in \sum_{\mathrm{f}}$. The cost function $c(\cdot)$ computes the cost in terms of Euclidean distance.

The following motion planning problems will be considered in the proposed algorithm:

**Problem 1 (Feasible path solution)** *Find a path $\sigma_\mathrm{f} : [0, s]$, if one exists, in obstacle-free space $X_\mathrm{free} \subset X$ such that $\sigma_\mathrm{f}(0) = x_\mathrm{init} \in X_\mathrm{free}$ and $\sigma_\mathrm{f}(s) \in X_\mathrm{goal}$. If no such path exists, report failure.*

**Problem 2 (Optimal path solution)** *Find an optimal path $\sigma^*_\mathrm{f} : [0, s]$ connecting $x_\mathrm{init}$ and $X_\mathrm{goal}$ in obstacle-free space $X_\mathrm{free} \subset X$, such that the cost of the path $\sigma^*_\mathrm{f}$ is minimum, i.e., $c(\sigma^*_\mathrm{f}) = \{\min_{\sigma_\mathrm{s}} c(\sigma_\mathrm{f}) : \sigma_\mathrm{f} \in \sum_\mathrm{f}\}$.*

**Problem 3 (Convergence to Optimal Solution)** *Find the optimal path $\sigma^*_\mathrm{f} : [0, s]$ in obstacle-free space $X_\mathrm{free} \subset X$ in the least possible time $t \in \mathbb{R}$.*

# 3 RRT* Algorithm

This section describes the RRT* algorithm [12]. Algorithm 1 is a slightly modified implementation of RRT*. In this version, improvements were made to the original algorithm in order to enhance the computational efficiency of RRT* by reducing the number of calls to the ObstacleFree procedure [29]. Following are some of the processes employed by RRT*:

**Algorithm 1:** RRT*($x_\mathrm{init}$)

```
V ← {x_init}; E ← ∅; T ← (V, E);
for i ← 0 to N do
    x_rand ← Sample(i)
    X_near ← NearVertices(x_rand, T)
    if X_near = ∅ then
        X_near ← NearestVertex(x_rand, T)
    L_s ← GetSortedList(x_rand, X_near)
    x_min ← ChooseBestParent(L_s)
    if x_min ≠ ∅ then
        T ← InsertVertex(x_rand, x_min, T)
        T ← RewireVertices(x_rand, L_s, E)
return T = (V, E)
```

*Random Sampling:* the Sample procedure returns an independent and uniformly distributed random sample from the obstacle free space, i.e., $x_\mathrm{rand} \in X_\mathrm{free}$.

*Collision Check:* the procedure ObstracleFree($\sigma$) checks whether the given path $\sigma : [0, 1]$ belongs to $X_\mathrm{free}$ or not. A true value is reported if $\sigma(s) \in X_\mathrm{free} \forall s[0, 1]$.

**Algorithm 2:** GetSortedList($x_\mathrm{rand}, X_\mathrm{near}$)

```
L_s ← ∅
for x' ∈ X_near do
    σ' ← Steer(x', x_rand)
    C' ← c(x_init, x') + c(x', x_rand)
    L_s ← (x', C', σ')
L_s ← SortList(L_s)
return L_s
```

*Near Vertices:* given a sample $x \in X$, the tree $T = (V, E)$ and the ball region $\mathfrak{B}_{x,r}$ of radius $r$ centered at $x$, the set of near vertices is defined as:

Near$(x, T, r) := \{v \in V : v \in \mathfrak{B}_{x,r}\} \mapsto X_\mathrm{near} \subseteq V$. More specifically, $X_\mathrm{near} = \{v \in V : d(x, v) \leq \gamma(logi/i)^{1/n}\}$ where $i$ is the number of vertices, $n$ represents the dimensions and $\gamma$ is a constant.

*Nearest Vertex:* As its name suggests, this procedure returns the nearest vertex in the tree from any given state $x \in X$. Given the tree $T = (V, E)$, the nearest vertex procedure can be defined as:

$$\mathrm{Nearest}(T, x) := \mathrm{argmin}_{v \in V} d(x, v) \mapsto x_\mathrm{min}.$$

*Getting Sorted List:* the procedure GetSortedList in Algorithm 2 constructs a tuple and returns it as the list $L_\mathrm{s}$. Each element of this list is a triplet of form $(x', c(\sigma), \sigma') \in L_\mathrm{s}$ where $x' \in X_\mathrm{near}$. The list $L_\mathrm{s}$ is sorted in the ascending value of the cost function.

*Steering:* the steering function utilised in this modified version of RRT* takes two states as an input and returns the straight trajectory connecting those two states. For example, for two given states $x_1 \in X$ and $x_2 \in X$, the path $\sigma : [0, 1]$ will be the path connecting these two states, i.e., $\sigma(0) = x_1$ and $\sigma(1) = x_2$. The steering is done from $x_1$ to $x_2$ in small, discrete steps and can be summarized as $\sigma(s') = (1 - s')x_1 + s'x_2; \forall s'[0, 1]$.

*Choosing Best Parent:* this procedure is used to search the list $L_\mathrm{s}$ for a state, $x_\mathrm{min} \in L_\mathrm{s}$ which provides the shortest, collision-free path $\sigma'$ from the initial state $x_\mathrm{init}$ to the random sample $x_\mathrm{rand}$. Alternatively, $\sigma'$ is the shortest collision-free path connecting the initial state $x_\mathrm{init}$ and the random sample $x_\mathrm{rand}$ via $x_{\min} \in L_{\mathrm{s}}$. Algorithm 3 outlines the implementation of this procedure.

**Algorithm 3:** ChooseBestParent($L_{\mathrm{s}}$)

**for** $(x', C', \sigma') \in L_{\mathrm{s}}$ **do**
    **if** ObstacleFree($\sigma'$) **then**
        **return** $x'$
**return** $\emptyset$

**Algorithm 4:** RewireVertices($x_{\mathrm{rand}}, L_{\mathrm{s}}, E$)

1 **for** $(x', C', \sigma') \in L_{\mathrm{s}}$ **do**
2     **if** $\big(c(x_{\mathrm{init}}, x_{\mathrm{rand}}) + c(x_{\mathrm{rand}}, x')\big) < c(x_{\mathrm{init}}, x')$ **then**
3         **if** ObstacleFree($\sigma_{\mathrm{new}}$) **then**
4             $x_{\mathrm{parent}} \leftarrow$ Parent$(E, x')$
5             $E \leftarrow (E \backslash \{(x_{\mathrm{parent}}, x')\}) \cup (\{x_{\mathrm{rand}}, x'\})$
6 **return** $E$

The RRT* algorithm provides asymptotic optimality. In reference to Algorithm 1, the RRT* algorithm after preliminary initialization process starts its iterative process by sampling the random sample $x_{\mathrm{rand}}$ from the given configuration space $X_{\mathrm{free}}$ (Line 3). After this, the RRT* finds the set of near vertices $X_{\mathrm{near}}$ from the tree lying inside the ball region centered at $x_{\mathrm{rand}}$. If the set of near vertices $X_{\mathrm{near}}$ computed by NearVertices procedure is empty, then the set $X_{\mathrm{near}}$ is filled by the NearestVertex procedure (Line 4-6). The populated set $X_{\mathrm{near}}$ is then sorted by the GetSortedList procedure to form a list of form $(x', c(\sigma), \sigma')$, arranged in ascending order of cost function $c(\sigma)$ (Line 7). The procedure ChooseBestParent iterates over the sorted list $L_{\mathrm{s}}$ (Line 8), returning the best parent vertex $x_{\min} \in X_{\mathrm{near}}$ through which $x_{\mathrm{init}}$ and $x_{\mathrm{rand}}$ can be connected in obstacle-free space. Once such a state is located, i.e, the best parent vertex $x_{\min}$ is no longer empty, $x_{\min}$ is inserted into the tree by making $x_{\mathrm{rand}}$ its child and then the rewiring step is executed (Line 9-11). Algorithm 4 presents the pseudocode of the rewiring process. Here, the algorithm examines each vertex $x' \in X_{\mathrm{near}}$ lying inside the ball region centered at $x_{\mathrm{rand}}$. If the cost of the path connecting $x_{\mathrm{init}}$ and $x'$ through $x_{\mathrm{rand}}$ is less than the existing cost of reaching $x'$ and if this path lies in obstacle-free space $X_{\mathrm{free}}$ (Algorithm 4 Line 1-3), then $x_{\mathrm{rand}}$ is made the parent of $x'$ (Algorithm 4 Line 4-5). If these conditions do not hold true, no change is made to the tree and the algorithm moves on to check the next vertex. This process is iteratively performed for every vertex $x'$ present in the sorted list $L_{\mathrm{s}}$.

**Algorithm 5:** B-RRT*($x_{\mathrm{init}}$,$x_{\mathrm{goal}}$)

$V \leftarrow \{x_{\mathrm{init}}, x_{\mathrm{goal}}\}; E \leftarrow \emptyset;$
  $T_{\mathrm{a}} \leftarrow (x_{\mathrm{init}}, E); T_{\mathrm{b}} \leftarrow (x_{\mathrm{goal}}, E);$
$\sigma_{\mathrm{best}} \leftarrow \infty$
**for** $i \leftarrow 0$ ***to*** $N$ **do**
    $x_{\mathrm{rand}} \leftarrow$ Sample$(i)$
    $x_{\mathrm{nearest}} \leftarrow$ NearestVertex$(x_{\mathrm{rand}}, T_{\mathrm{a}})$
    $x_{\mathrm{new}} \leftarrow$ Extend$(x_{\mathrm{nearest}}, x_{\mathrm{rand}})$
    $X_{\mathrm{near}} \leftarrow$ NearVertices$(x_{\mathrm{new}}, T_{\mathrm{a}})$
    $L_{\mathrm{s}} \leftarrow$ GetSortedList$(x_{\mathrm{new}}, X_{\mathrm{near}})$
    $x_{\min} \leftarrow$ ChooseBestParent$(L_{\mathrm{s}})$
    **if** $x_{\min} \neq \emptyset$ **then**
        $T \leftarrow$ InsertVertex$(x_{\mathrm{new}}, x_{\min}, T_{\mathrm{a}})$
        $T \leftarrow$ RewireVertices$(x_{\mathrm{new}}, L_{\mathrm{s}}, E)$
    $x_{\mathrm{conn}} \leftarrow$ NearestVertex$(x_{\mathrm{new}}, T_{\mathrm{b}})$
    $\sigma_{\mathrm{new}} \leftarrow$ Connect$(x_{\mathrm{new}}, x_{\mathrm{conn}}, T_{\mathrm{b}})$
    **if** $\sigma_{\mathrm{new}} \neq \emptyset \&\& c(\sigma_{\mathrm{new}}) < c(\sigma_{\mathrm{best}})$ **then**
        $\sigma_{\mathrm{best}} \leftarrow \sigma_{\mathrm{new}}$
    SwapTrees$(T_{\mathrm{a}}, T_{\mathrm{b}})$
**return** $T_{\mathrm{a}}, T_{\mathrm{b}} = (V, E)$

# 4 B-RRT* algorithm

This section explains the implementation of Bidirectional-RRT*(B-RRT*) [11]. Algorithm 5 outlines the implementation of B-RRT*; extra procedures employed by B-RRT* are explained below while the rest are exactly the same as they were for RRT*.

*Extend:* given two nodes $x_1, x_2 \in X$, the Extend$(x_1, x_2)$ procedure returns a new node $x_{\text{new}} \in \mathbb{R}^n$ such that $x_{\text{new}}$ is more closer to $x_2$ than $x_1$ in the direction from $x_1$ to $x_2$.

*Connect:* connect heuristic employed by B-RRT* is slight variation of greedy RRT-Connect heuristic[20]. Algorithm 6 outlines the implementation of Connect heuristic of B-RRT*. Typical RRT* iteration is performed on the input nodes $x_1, x_2$, where $x_1$ plays the role of $x_{\text{rand}}$ while the set of near vertices is computed from the other tree (Line 1-2). After computing a set of near vertices from tree b, the procedure GetSortedList (explained in previous section) is executed, and the best vertex is selected from the sorted list such that it provides collision-free low-cost connection between the trees $T_{\text{a}}, T_{\text{b}}$. Finally, this procedure ends by generating and returning the end-to-end feasible path solution, connecting $x_{\text{init}}$ and $X_{\text{goal}}$.

**Algorithm 6:** Connect($x_1$,$x_2$,$T_{\text{b}}$)

1 $x_{\text{new}} \leftarrow$ Extend$(x_2, x_1)$
2 $X_{\text{near}} \leftarrow$ NearVertices$(x_{\text{new}}, T_{\text{b}})$
3 $L_{\text{s}} \leftarrow$ GetSortedList$(x_1, X_{\text{near}})$
4 $x_{\text{min}} \leftarrow$ ChooseBestParent$(L_{\text{s}})$
5 **if** $x_{\text{min}} \neq \emptyset$ **then**
6     $E \leftarrow E \cup ((x_{\text{min}}, x_1))$
7     $\sigma_{\text{f}} \leftarrow$ GeneratePath$(x_{\text{min}}, x_1)$
8     **return** $\sigma_{\text{f}}$
9 **return** NULL

In reference to Algorithm 5, the B-RRT* works in exactly the same manner as the original RRT* algorithm in its initial phases i.e., it starts with sampling of the configuration space $X_{\text{free}}$, then various operations (just like RRT*) are performed on this random sample $x_{\text{rand}}$ (Line 4-12 ). After successful insertion of random sample into the tree under operation (Line 11-12), the algorithm computes nearest vertex $x_{\text{conn}}$ from $x_{\text{new}}$ in the tree $T_{\text{b}}$, and then executes the connect procedure for the connection of two trees (Line 13-14). If the attempt to make connection is successful, the collision-free path $\sigma_{\text{new}}$ connecting $x_{\text{init}}$ and $X_{\text{goal}}$ is returned by the connect function. The cost of this $\sigma_{\text{new}}$ is then compared with the previously computed path $\sigma_{\text{best}}$. If the cost of $\sigma_{\text{new}}$ is less than the cost of $\sigma_{\text{best}}$, then $\sigma_{\text{best}}$ is overwritten by $\sigma_{\text{new}}$ (Line 15-16). Finally, the iteration ends by swapping the trees (Line 17) and in the next iteration the same procedure is executed on the other tree.

# 5 IB-RRT* algorithm

This section presents our proposed IB-RRT* algorithm[1]. IB-RRT* is specifically designed for motion

**Algorithm 7:** IB-RRT* $(x^{\text{a}}_{\text{init}}, x^{\text{b}}_{\text{init}})$

$V_{\text{a}} \leftarrow \{x^{\text{a}}_{\text{init}}\}; E_{\text{a}} \leftarrow \emptyset; T_{\text{a}} \leftarrow (V_{\text{a}}, E_{\text{a}})$
$V_{\text{b}} \leftarrow \{x^{\text{b}}_{\text{init}}\}; E_{\text{b}} \leftarrow \emptyset; T_{\text{b}} \leftarrow (V_{\text{b}}, E_{\text{b}})$
$\sigma_{\text{f}} \leftarrow \infty; E \leftarrow \emptyset$
Connection $\leftarrow$ True
**for** $i \leftarrow 0$ ***to*** $N$ **do**
    $x_{\text{rand}} \leftarrow$ Sample$(i)$
    $\{X^{\text{a}}_{\text{near}}, X^{\text{b}}_{\text{near}}\} \leftarrow$ NearVertices$(x_{\text{rand}}, T_{\text{a}}, T_{\text{b}})$
    **if** $X^{\text{a}}_{\text{near}} = \emptyset$ & & $X^{\text{b}}_{\text{near}} = \emptyset$ **then**
        $\{X^{\text{a}}_{\text{near}}, X^{\text{b}}_{\text{near}}\} \leftarrow$ NearestVertex$(x_{\text{rand}}, T_{\text{a}}, T_{\text{b}})$
        Connection $\leftarrow$ False
    $L^{\text{a}}_{\text{s}} \leftarrow$ GetSortedList$(x_{\text{rand}}, X^{\text{a}}_{\text{near}})$
    $L^{\text{b}}_{\text{s}} \leftarrow$ GetSortedList$(x_{\text{rand}}, X^{\text{b}}_{\text{near}})$
    $\{x_{\text{min}}, \text{flag}, \sigma_{\text{f}}\} \leftarrow$ GetBestTreeParent$(L^{\text{a}}_{\text{s}}, L^{\text{b}}_{\text{s}}, \text{Connection})$
    **if** (flag) **then**
        $T_{\text{a}} \leftarrow$ InsertVertex$(x_{\text{rand}}, x_{\text{min}}, T_{\text{a}})$
        $T_{\text{a}} \leftarrow$ RewireVertices$(x_{\text{rand}}, L_{\text{s}}, E_{\text{a}})$
    **else**
        $T_{\text{b}} \leftarrow$ InsertVertex$(x_{\text{rand}}, x_{\text{min}}, T_{\text{b}})$
        $T_{\text{b}} \leftarrow$ RewireVertices$(x_{\text{rand}}, L_{\text{s}}, E_{\text{b}})$
$E \leftarrow E_{\text{a}} \cup E_{\text{b}}$
$V \leftarrow V_{\text{a}} \cup V_{\text{b}}$
**return** $(\{T_{\text{a}}, T_{\text{b}}\} = V, E)$

[1]The source code is available at :github.com/ahq1993

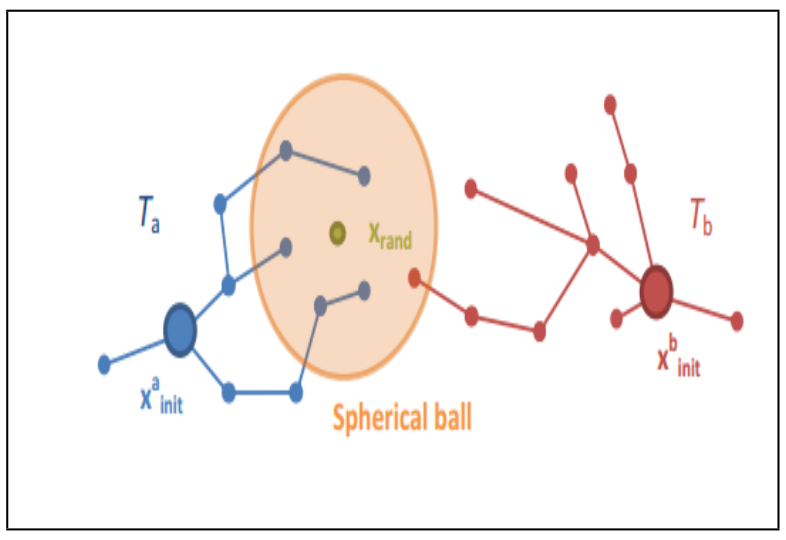

(a) Computing set of near nodes

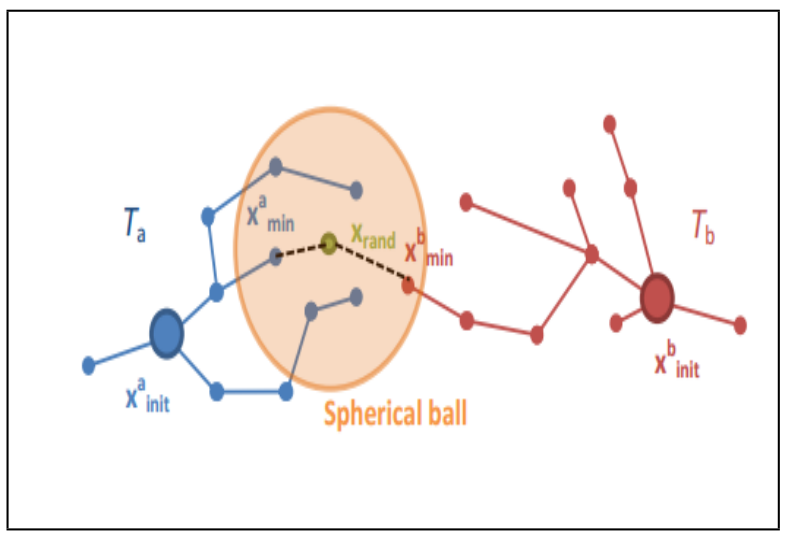

(b) Choosing best parent

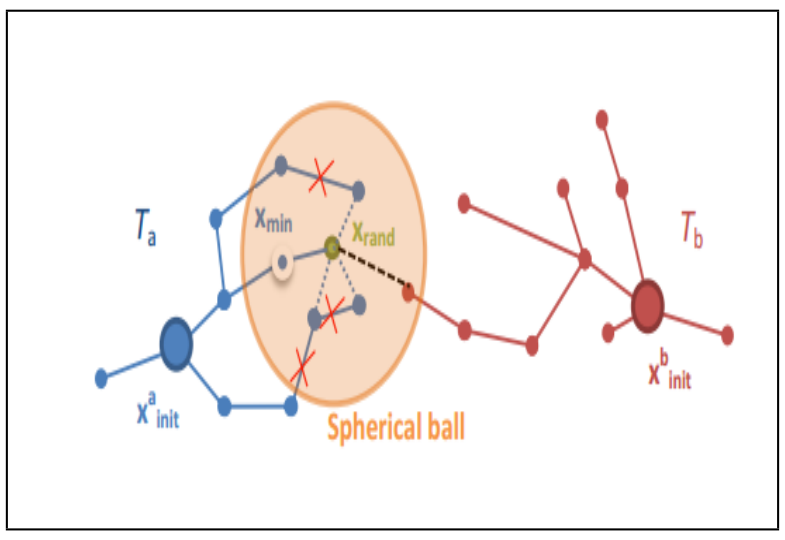

(c) Choosing best parent vertex and rewiring

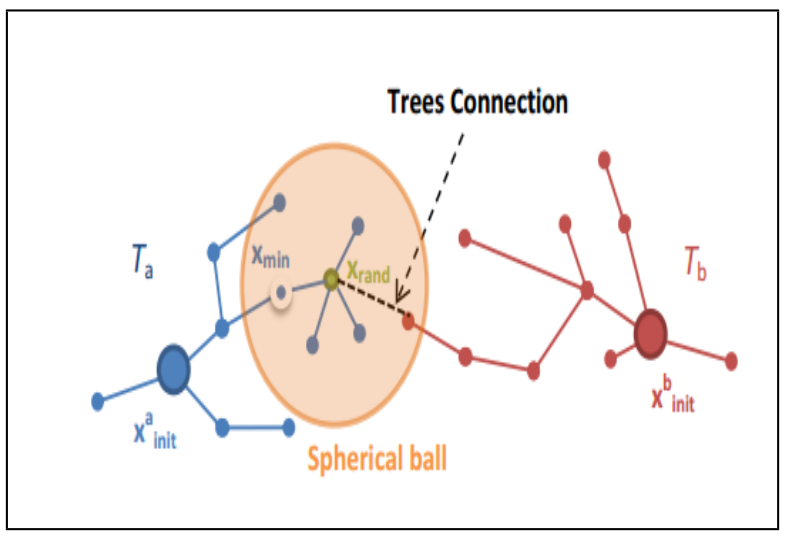

(d) Connecting trees

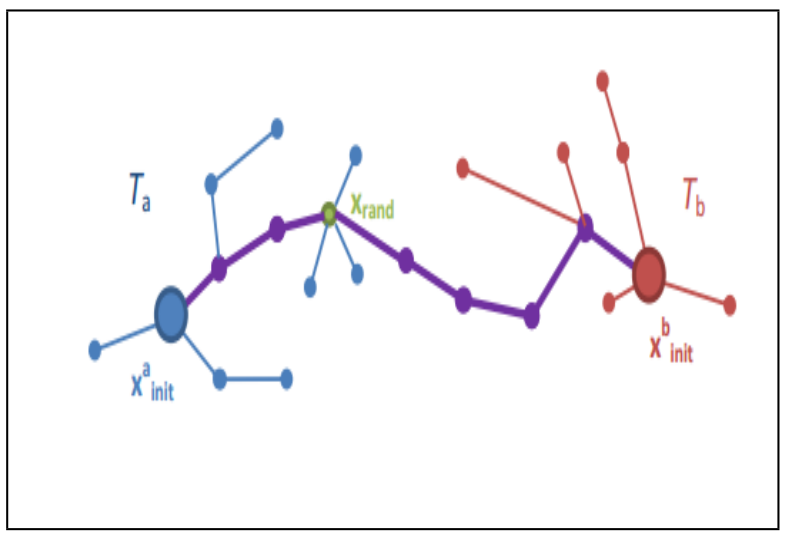

(e) end-to-end path

Figure 1: Intelligent Bidirectional Trees

planning in complex cluttered environments where exploration of configuration space is difficult. Let the sets of near vertices from tree $T_{\mathrm{a}}$ and $T_{\mathrm{b}}$ be denoted by $X^{\mathrm{a}}_{\mathrm{near}}$ and $X^{\mathrm{b}}_{\mathrm{near}}$, respectively. The path connecting $x^{\mathrm{a}}_{\mathrm{init}}$ and $x_{\mathrm{rand}}$ is denoted by $\sigma'_{\mathrm{a}} : [0, s_{\mathrm{a}}]$, while the path connecting $x^{\mathrm{b}}_{\mathrm{init}}$ and $x_{\mathrm{rand}}$ is denoted by $\sigma'_{\mathrm{b}} : [0, s_{\mathrm{b}}]$. Algorithm 7 outlines the implementation of IB-RRT*. In Algorithm 7, the boolean variable Connection represents the feasibility of connecting the two trees while the boolean variable flag indicates the best selected tree. IB-RRT* builds the bidirectional trees incrementally. It starts by picking a random sample $x_{\mathrm{rand}}$ from the obstacle-free configuration space $X_{\mathrm{free}}$ i.e., $x_{\mathrm{rand}} \in X_{\mathrm{free}}$ (Line 6). It then populates the set of near vertices $X^{\mathrm{a}}_{\mathrm{near}}, X^{\mathrm{b}}_{\mathrm{near}}$ for both trees using the NearVertices procedure (Line 7). It should be noted that a ball region centered at $x_{\mathrm{rand}}$ of radius $r$ is formed and the sets of the near vertices from both trees are computed i.e., $X^{\mathrm{a}}_{\mathrm{near}} := \{v \in V_{\mathrm{a}} : v \in \mathfrak{B}_{x_{\mathrm{rand}},r}\}$ and $X^{\mathrm{b}}_{\mathrm{near}} := \{v \in V_{\mathrm{b}} : v \in \mathfrak{B}_{x_{\mathrm{rand}},r}\}$, as shown in figure 1(a). $X^{\mathrm{a}}_{\mathrm{near}}$ and $X^{\mathrm{b}}_{\mathrm{near}}$ now contain near vertices of $x_{\mathrm{rand}}$ from trees $T_a$ and $T_b$, respectively. In case of both sets of near vertices being found empty, these sets are filled with the closest vertex from their respective trees instead, i.e, the vertex on their respective tree which lies closest to the random sample(Line 8-9). Both sets are then sorted by the GetSortedList procedure (Line 11-12) outlined in Algorithm 2. Once the list is in ascending order, the random vertex $x_{\mathrm{rand}}$ is inserted in the best selected tree (Line 13-19). The procedure BestSelectedTree (explained later in this section) returns the nearest vertex on best selected tree which is eligible to become parent of the random sample.

Additional features included in IB-RRT* are explained below. The rest of the procedures employed by our algorithm are the same as those outlined in the previous section.

*Selecting Best Tree Parent:* the operation GetBestTreeParent replaces the ChooseBestParent

procedure of the RRT* algorithm. The implementation of this procedure is outlined in Algorithm 8. Initially the best parent vertex from both the trees is calculated, as shown in figure 1(b). This process (Line 2-9) is similar to ChooseBestParent procedure outlined in algorithm 3. The best selected triplets from each tree $T_{\mathrm{a}}$ and $T_{\mathrm{b}}$ are assigned to $\{x^{\mathrm{a}}_{\mathrm{min}}, C^{\mathrm{a}}_{\mathrm{min}}, \sigma_{\mathrm{a}}\} \in L^{\mathrm{a}}_{\mathrm{s}}$ and $\{x^{\mathrm{b}}_{\mathrm{min}}, C^{\mathrm{b}}_{\mathrm{min}}, \sigma_{\mathrm{b}}\} \in L^{\mathrm{b}}_{\mathrm{s}}$ respectively. Following this, the GetBestTreeParent procedure selects the best tree from amongst $T_{\mathrm{a}}$ and $T_{\mathrm{b}}$ on the basis of costs $C^{\mathrm{a}}_{\mathrm{min}}$ and $C^{\mathrm{b}}_{\mathrm{min}}$ associated with each best selected triplet. The best selected vertex of the best selected tree, i.e. either $x^{\mathrm{a}}_{\mathrm{min}}$ or $x^{\mathrm{b}}_{\mathrm{min}}$, is then assigned to $x_{\mathrm{min}}$ for the insertion process. For the scenario depicted in Figure 1, tree $T_{\mathrm{a}}$ is selected as the best tree and therefore $x^{\mathrm{a}}_{\mathrm{min}}$ is assigned to $x_{\mathrm{min}}$ as shown in the figure 1c. The boolean variable flag indicates which tree has been selected as the best tree for any single iteration (Line 10-14). The algorithm then attempts to connect the bidirectional trees (Line 15) on the basis of the boolean variable connection. This is explained further on in the paper. The GetBestTreeParent procedure concludes by returning the best vertex $x_{\mathrm{min}}$, the boolean flag and, if it exists, the final path $\sigma_{\mathrm{f}}$ connecting the initial state to the goal region.

*Bidirectional Trees Connection:* Algorithm 9 gives the pseudocode of the procedure ConnectTrees. Given collision-free paths $\sigma_{\mathrm{a}} : [0, s_{\mathrm{a}}]$ and $\sigma_{\mathrm{b}} : [0, \mathrm{b}]$, where $\sigma_{\mathrm{a}}(0) = x^{\mathrm{a}}_{\mathrm{init}}$, $\sigma_{\mathrm{b}}(0) = x^{\mathrm{b}}_{\mathrm{init}}$ and $\sigma_{\mathrm{a}}(s_{\mathrm{a}}) = \sigma_{\mathrm{b}}(s_{\mathrm{b}}) = x_{\mathrm{rand}}$. This procedure updates the end-to-end collision-free path $\sigma_{\mathrm{f}} : [0, s]$ connecting $\sigma_{\mathrm{f}}(0) = x^{\mathrm{a}}_{\mathrm{init}}$ and $\sigma_{\mathrm{f}}(s) = x^{\mathrm{b}}_{\mathrm{init}}$ if the cost of concatenated paths, $c(\sigma_{\mathrm{a}}|\sigma_{\mathrm{b}})$, is found to be less than the cost of the existing end-to-end path $c(\sigma_{\mathrm{f}})$ (Line 1-2). Connection between the trees is only successful if the boolean variable connection is true. As mentioned in previous explanation, the occurrence of empty sets for both $X^{\mathrm{a}}_{\mathrm{near}}$ and $X^{\mathrm{b}}_{\mathrm{near}}$ causes the procedure NearestVertex to be called (Algorithm 7, Line 7-8). The NearestVertex changes the boolean variable connection to false. Therefore, the boolean connection is true only when the procedure NearVertices populates both sets. This implies that the two trees are connected if ball of region centered at $x_{\mathrm{rand}}$ contains near vertices from both trees $T_{\mathrm{a}}$ and $T_{\mathrm{b}}$. Hence, unlike the connect heuristic [20], the IB-RRT* is not greedy since the connection is only made inside the ball region as shown in the figure 1(d). Finally the tree connection generates end-to-end global path, as shown in figure 1(e).

**Algorithm 8:** GetBestTreeParent($L^{\mathrm{a}}_{\mathrm{s}}, L^{\mathrm{b}}_{\mathrm{s}}$, Connection)

```
1  flag ← True
2  for (x'_a, C'_a, σ'_a) ∈ L^a_s do
3      if ObstacleFree(σ'_a) then
4          {x^a_min, C^a_min, σ_a} ← {x'_a, C'_a, σ'_a}
5          Break
6  for (x'_b, C'_b, σ'_b) ∈ L^b_s do
7      if ObstacleFree(σ'_b) then
8          {x^b_min, C^b_min, σ_b} ← {x'_b, C'_b, σ'_b}
9          Break
10 if (C^a_min ≤ C^b_min) then
11     x_min ← x^a_min
12 else if (C^b_min < C^a_min) then
13     x_min ← x^b_min
14     flag ← False
15 if Connection then
16     σ_f ← ConnectTrees(σ_a, σ_b)
17 return {x_min, flag, σ_f}
```

**Algorithm 9:** ConnectTrees($\sigma_{\mathrm{a}}, \sigma_{\mathrm{b}}$)

```
1  if c(σ_f) < c(σ_a|σ_b) then
2      σ_f ← σ_a|σ_b
3  return σ_f
```

# 6 Analysis

## 6.1 Probabilistic Completeness

In any configuration space, an algorithm is said to be *Probabilistically Complete* if the probability of finding a path solution, if ones exist, approaches one as the number of samples taken from the configuration space reaches infinity. It is known that RRT is a probabilistically complete algorithm, as its optimal variant RRT* [13]. Since our proposed IB-RRT* algorithm performs the random sampling function exactly like the aforementioned algorithms and is merely a bidirectional version of RRT* with intelligent sample insertion, it can be reasonably proffered that it also inherits the probabilistic completeness property of RRT*.

## 6.2 Asymptotic Optimality

It is known that RRT and its variant RRT-Connect do not ensure optimality even if the number of iterations are increased to infinity [13]. However, RRT* is an optimal variant of RRT, ensuring almost-sure convergence to an optimal solution [12]. As explained earlier, IB-RRT* attempts to connect both trees, $T_{\rm a} = (V_{\rm a}, E_{\rm a})$ and $T_{\rm b} = (V_{\rm b}, E_{\rm b})$, in every iteration. A random sample $x_{\rm rand}$ is used as a point of connection between the two trees (shown in figure 1(d)) if the ball region centered at $x_{\rm rand}$ is found to contain near vertices from both the trees, i.e., $v_{\rm a} \in V_{\rm a} : v_{\rm a} \in \mathfrak{B}_{x_{\rm rand},r}$ and $v_{\rm b} \in V_{\rm b} : v_{\rm b} \in \mathfrak{B}_{x_{\rm rand},r}$. A similar procedure is employed by the RRT* algorithm [12] to connect the random sample with its chosen parent. Since there is no extra connection heuristic required for connection of the two trees and the two trees are generated exactly as the tree generated in the original RRT* algorithm, it can be reasonably proposed that the IB-RRT* algorithm inherits the asymptotic optimality property of RRT*.

## 6.3 Rapid Convergence to Optimal Path

This section provides proof of IB-RRT*'s rapid convergence to the optimal solution and that this algorithm provides faster convergence rates as compared to state of the art algorithms RRT* and B-RRT*. For simplicity, the following assumptions are are made:

**Assumption 1 (Uniform Sampling)** *The sampling operation take samples from a configuration space $X$ such that the samples are continuously distributed.*

**Assumption 2 (Cluttered Configuration Space)** *The configuration space $X$ is cluttered such that the tree initially grows near its initial state $x_{\rm init}$ and then incrementally grows towards the unsearched configuration space.*

**Assumption 3 (Uniformity and Additivity of Cost Function)** *For a given set of path functions, the cost function must satisfy: $c(\sigma_1) \leq c(\sigma_1|\sigma_2) : c(\sigma_1|\sigma_2) = c(\sigma_1) + c(\sigma_2) \forall \sigma_1, \sigma_2 \in \sum_{\rm free}$.*

Assumption 1 ensures that the sampling operation is not biased or goal directed. Biasing the samples for rapid convergence to the optimal solution is computationally inefficient, specifically in higher dimension configuration spaces [13]. Assumption 2 states that the environment contains obstacles that hinder the expansion of the two trees in the configuration space. Finally, Assumption 3 simply asserts that the longer path has a higher cost than the shorter one.
As mentioned in section II, $\sum_{\rm free}$ denote the set of all collision free paths in the tree $T = (V, E)$. Let $\sigma_i, \sigma_i' \in \sum_{\rm free}$ such that $\sigma_i'$ is closest to $\sigma_i$ in terms of Euclidean distance function. The following Lemma states that any sampling-based algorithm can provide almost-sure convergence to the optimal path solution if distance variation $\|\sigma_i' - \sigma_i\|$ approaches zero as the number of iterations approaches infinity.

**Lemma 1 ([13])** *A sampling-based algorithm ensures asymptotic optimality, such that*
$\mathbb{P}(\lim_{i\to\infty} \|\sigma_i' - \sigma_i\| = 0; \forall \sigma_i, \sigma_i' \in \sum_{\rm free}) = 1$.

With lemma 1 following corollary immediate.

**Corollary 1** *By increasing the number of path variations minimized per iteration, the algorithm can greatly improve its rate of convergence to an optimal solution.*

Given a tree $T = (V, E)$, random configuration state $x \in X_{\rm free}$ and a set of near vertices $X_{\rm near}$ inside a ball region $\mathfrak{B}_{x,r}$ centered at $x$, the intensity of near vertices around $x$, denoted by $\mathrm{J}_x$, can be defined as:

$J_x := \{\text{card}(X_{\text{near}})/\mu(\mathfrak{B}_{x,r}) : X_{\text{near}}|x = \mathfrak{B}_{x,r} \cap V\}$.

Regarding the intensity of near vertices Lemma 2 is stated as follows:

**Lemma 2** *If Assumptions 1,2,3 hold, then Intensity $J_x$ is higher in the regions closer to the point of generation of the tree.*

**Sketch of Proof:** Let $\epsilon \in \mathbb{R}_+$. $X_{\text{free}}$ is obstacle-free configuration space. $X_{\text{free}}$ is searched for the set of near neighbors $X_{\text{near}}$ that lie inside a ball region $\mathfrak{B}_{x,r}$ of radius $r > 0$ centered at the random state $x \in X_{\text{free}}$. Any state $x' \in X_{\text{near}}$ can become the parent of $x_{\text{rand}}$, if it provides a lower cost path connecting $x_{\text{rand}}$ to $x_{\text{init}}$ than the one provided by all other vertices in $X_{\text{near}}$. This implies that $\|x - x'\| = \epsilon$, where $\epsilon < r = \gamma(logi/i)^{1/n}$. This ensures that the growth of the algorithm presented in this paper is incremental as the tree grows in small incremental distances $\epsilon$. For the incremental or wavefront expansion of the trees it is proven that the regions near the point of generation of trees are more dense [23]. Therefore, there is a high probability of having high cardinality of set $X_{\text{near}}$, if the random state lies closer to the point of generation of tree.

Hence, with corollary 1 and Lemma 2 holds the following theorem stating effectiveness of IB-RRT* is given below.

**Theorem 1** *IB-RRT* algorithm converges to optimal solution more quickly as compared to RRT* and B-RRT*.*

**Sketch of Proof:** Given a random configuration $x_{\text{rand}}$ and minimum cost path functions $\sigma_{\text{a}}[0, s_{\text{a}}] := \{\sigma_{\text{a}}(0) = x^{\text{a}}_{\text{init}}, \sigma_{\text{a}}(s_{\text{a}}) = x_{\text{rand}}\}$ and $\sigma_{\text{b}}[0, s_{\text{b}}] := \{\sigma_{\text{b}}(0) = x^{\text{b}}_{\text{init}}, \sigma_{\text{b}}(s_{\text{b}}) = x_{\text{rand}}\}$, then insertion process of IB-RRT* can be summarized as $\{x_{\text{rand}} \in V_{\text{a}} : c(\sigma_{\text{a}}) \leq c(\sigma_{\text{b}})$ otherwise $x_{\text{rand}} \in V_{\text{b}} : c(\sigma_{\text{b}}) < c(\sigma_{\text{a}})\}$ (Algorithm 8). As the random sample $x_{\text{rand}}$ is always inserted into the tree whose initial state is closer to $x_{\text{rand}}$, this ensures that the sample is inserted into a region in the configuration space where the intensity of near vertices $J_x$ is high. Since the rewiring process explained earlier tries to minimize the distance variation $\|\sigma'_i - \sigma_i\|$ between any two closest paths in each tree. This is done by checking viability of the random sample $x_{\text{rand}}$ as the parent of each vertex in the set $X_{\text{near}}$. If the cost to reach a particular vertex $x'$ in the near set $X_{\text{near}}$ through random sample $x_{\text{rand}}$ is lower then the existing cost, then $x_{\text{rand}}$ becomes the parent of that particular vertex $x' \in X_{\text{near}}$. Hence, IB-RRT* inserts the sample into high intensity regions $J_x$, maximizing the rewiring process per iteration. This step allows rapid convergence to optimal solution and serves as evidence that IB-RRT* provides better convergence rates than both RRT* and B-RRT* algorithms. Furthermore, trees connection heuristic employed by B-RRT* [11] is partially greedy, similar to the the connect heuristic [20]. It has already been proved that if the bidirectional version of RRT* uses purely RRT-Connect heuristic [20] for the connection of two trees, it is no longer asymptotically optimal [11]. This happens because when only the connect heuristic [20] is used, an edge originating from $T_{\text{a}}$ for example, tries to reach the closest vertex on $T_{\text{b}}$. This implies that near vertices present inside the ball region are never considered for best parent selection. B-RRT* does eventually converge to an optimal solution but the convergence process is slowed down due to its partially greedy characteristic. Nevertheless, compared to RRT*, the B-RRT* has faster convergence rate due to its generation of two trees. However, in comparison to IB-RRT*, it has significantly less convergence rate. This is later on evident from the experimental results as well.

## 6.4 Computational Complexity

This section compares computational complexities of IB-RRT* with complexities of RRT* and the bidirectional version of RRT. Let $S_i^{\text{RRT}^*}$ and $S_i^{\text{BiRRT}}$ denotes the number of processes executed per iteration by RRT* and bidirectional-RRT (BiRRT), respectively. Let $S_i^{\text{Ours}}$ denote the number of processes executed by IB-RRT*. Theorem 2 and 3 propose that the running time of all processes executed per iteration by IB-RRT* is a constant times higher than both RRT* and BiRRT.

| Environment | Algorithm | $i_{\min}$ | $i_{\max}$ | $i_{\text{avg}}$ | $t_{\min}(s)$ | $t_{\max}(s)$ | $t_{\text{avg}}(s)$ | $C$ | Fail |
|---|---|---|---|---|---|---|---|---|---|
| 2D-Cluttered (A) (figures 2 & 3) | IB-RRT* | 159861 | 181521 | 162809 | 37.9 | 42.3 | 39.6 | 93.1 | 0 |
| | B-RRT* | 438785 | 463526 | 458328 | 90.8 | 97.8 | 95.3 | 93.1 | 3 |
| | RRT* | - | - | - | - | - | - | - | 50 |
| 2D-Cluttered (B) (figures 4 & 5) | IB-RRT* | 78652 | 141936 | 95192 | 18.9 | 33.4 | 23.3 | 69.4 | 0 |
| | B-RRT* | 261716 | 375341 | 286721 | 55.2 | 78.2 | 60.3 | 69.4 | 0 |
| | RRT* | 981431 | 1219628 | 1059268 | 168.9 | 210.9 | 183.4 | 69.4 | 7 |
| 3D-Multiple Barriers (figure 8) | IB-RRT* | 193593 | 218586 | 204321 | 45.8 | 53.6 | 47.8 | 81.9 | 0 |
| | B-RRT* | 810581 | 853248 | 838692 | 167.3 | 178.1 | 176.3 | 81.9 | 4 |
| | RRT* | 1941613 | 1978581 | 1961825 | 329.7 | 337.4 | 332.2 | 81.9 | 11 |
| 3D- Narrow Passages (figure 7) | IB-RRT* | 29651 | 34239 | 32361 | 6.8 | 8.2 | 7.8 | 69.8 | 1 |
| | B-RRT* | 46971 | 57891 | 54916 | 9.8 | 12.4 | 11.3 | 69.8 | 5 |
| | RRT* | 163872 | 168494 | 165627 | 27.2 | 29.4 | 28.6 | 69.8 | 8 |
| 3D-Maze (figure 6) | IB-RRT* | 148786 | 171543 | 168932 | 33.9 | 41.5 | 39.8 | 299.2 | 3 |
| | B-RRT* | 753861 | 764926 | 758438 | 155.3 | 161.2 | 159.7 | 299.2 | 7 |
| | RRT* | 2174180 | 2189742 | 2184761 | 368 | 372 | 371 | 299.2 | 16 |

Table 1: Experimental results for computing optimal path solution.

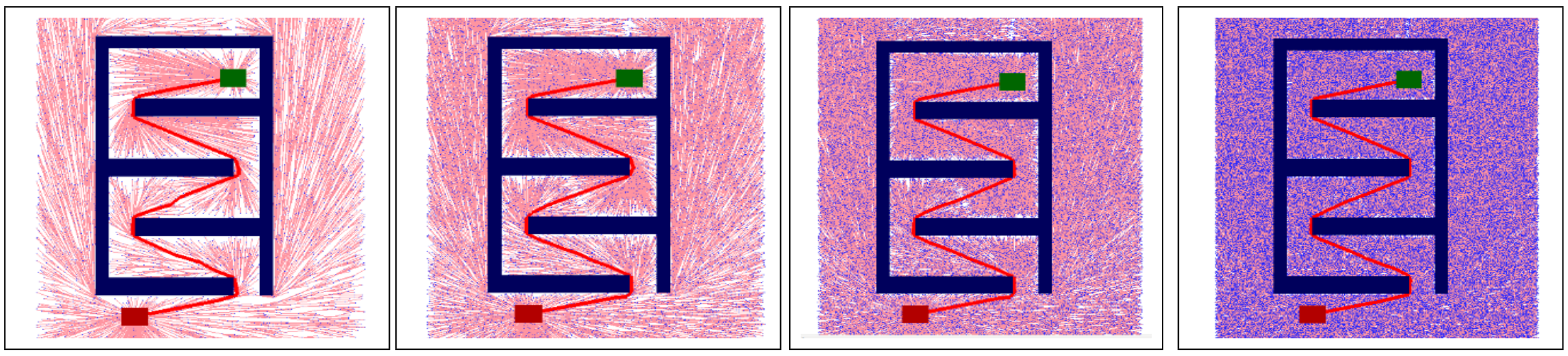

(a) i=7988,t=1.8s,C=95 (b) i=22164,t=4.9s,C=93.9 (c) i=79187,t=18.8s,C=93.4 (d) i=161437,t=40s,C=93.1

Figure 2: IB-RRT* performance in 2-D Environment (A)

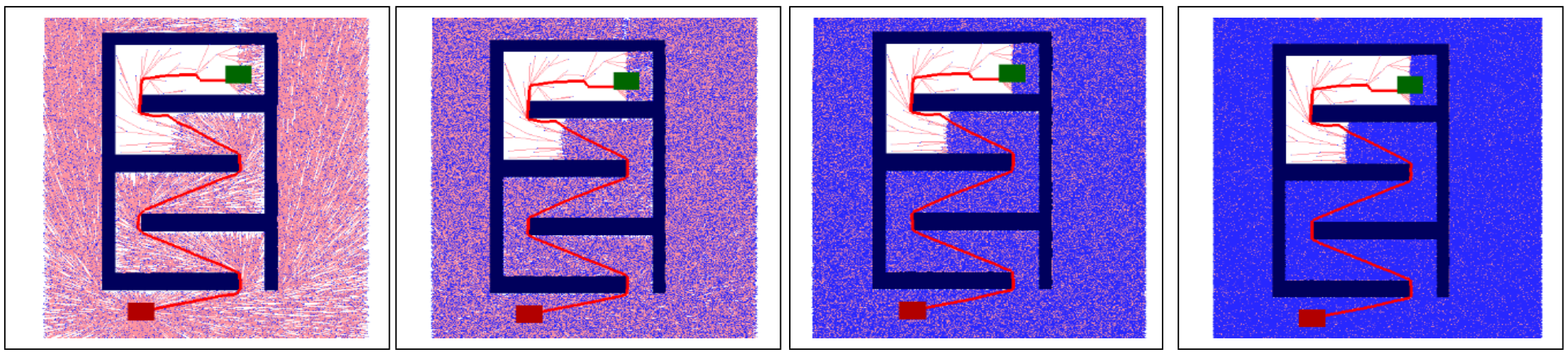

(a) i=102016,t=17s,C=97.0 (b) i=510982,t=84s,C=96.5 (c) i=1532191,t= 255s,C=96.3 (d) i=3192640,t=510s,C=96.0

Figure 3: RRT* performance in 2-D Environment (A)

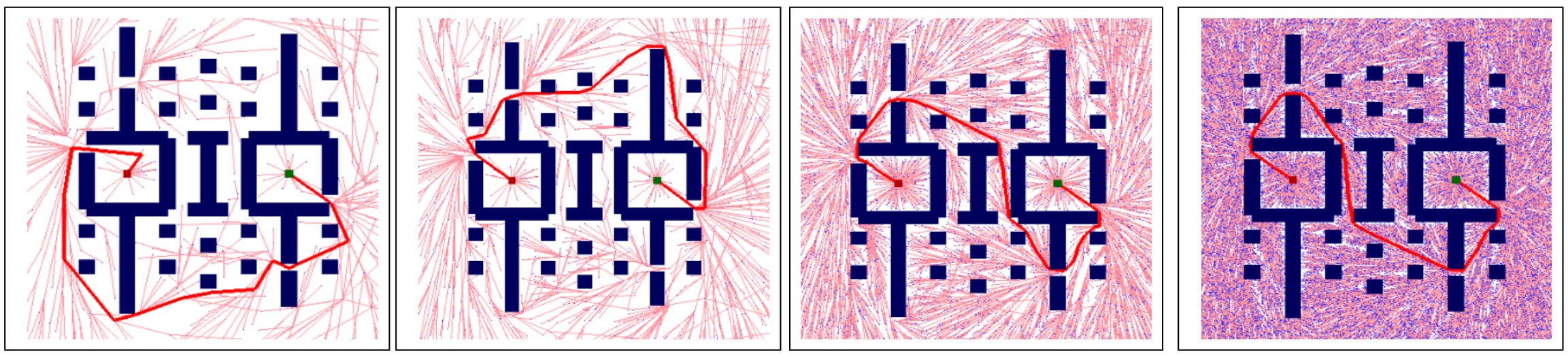

(a) i=1812,t=0.42s,C=83.4 (b) i=5026,t=1.22s,C=75.4 (c) i=30641,t=7.8s,C=70.8 (d) i=94865,t=21.6s,C=69.4

Figure 4: IB-RRT* performance in 2-D environment (B)

**Theorem 2** *The computational ratio of IB-RRT* and RRT* is such that there exists a constant* $\phi_1$ *i.e.,*

$$\lim_{i\to\infty} \mathbb{E}\left[\frac{S_i^{\text{Ours}}}{S_i^{\text{RRT}^*}}\right] \leq \phi_1.$$

**Theorem 3** *The computational ratio of IB-RRT* and BiRRT is such that there exists a constant* $\phi_2$ *i.e.,*

$$\lim_{i\to\infty} \mathbb{E}\left[\frac{S_i^{\text{Ours}}}{S_i^{\text{BiRRT}}}\right] \leq \phi_2.$$

Similar to RRT*, our proposed algorithm calls the procedures Sample and RewireVertices exactly once. The procedure of choosing best parent in RRT* is replaced by FindBestTree in the IB-RRT* algorithm, which also includes the ConnectTrees procedure. As explained earlier, the ConnectTrees function has negligible computational overhead since it merely concatenates two paths. In Intelligent Bidirectional-RRT* (IB-RRT*), for every iteration the procedures NearestVertex and NearVertices are executed for both trees $T_a$ and $T_b$. It has already been proved that both procedures have to run in *logi* expected time [12]. Furthermore, while IB-RRT* makes its best effort to increase the cardinality of the near vertices, the number of near vertices per tree returned by the procedure NearVertices cannot exceed a constant number [12]. Hence, it can be concluded that the execution of NearestVertex and NearVertices procedures on both trees per iteration adds up a constant computational complexity overhead as compared to RRT*. Hence, it can be concluded that IB-RRT* has the same computational complexity as RRT*. The proof for theorem 3 is exactly the same to one provided for the computational ratio of RRT* and RRT in [12].

# 7 Experimental Results

This section presents simulations performed on a 2.4GHz Intel corei5 processor with 4GB RAM. Here, performance results of our IB-RRT* algorithm are compared with RRT* and B-RRT*. Since exploration of the configuration space by B-RRT* after a large number of iterations is similar to that of IB-RRT*, snap shots presented here only depict the IB-RRT* and RRT* algorithms. This is to better demonstrate the difference between the expansion of trees of the two types of algorithms. For proper comparison, experimental conditions and size of the configuration space were kept constant for all algorithms. Since randomized sampling-based algorithms exhibit large variations in results, the algorithms were run up to 50 times with different seed values for each type of environment. Maximum, minimum and average number of iterations $i$ as well as time $t$ utilized by each algorithm to reach the optimal path solution is presented in the Table 1. To restrain the computational time within reasonable limits, the maximum limit for the number of tree nodes was kept at 5 million. The column fail in the table denotes the number of runs for which the corresponding algorithm failed to find an optimal path solution within node limits when executed with different seed values for random function. Although, algorithms were able to deter-mine feasible path solution, this is still considered as a failure, since the table provides comparison for the determination of an optimal path solution only. Figures 2 and 3 illustrate the trees maintained by IB-RRT* and RRT* respectively at different numbers of iterations. The cost C of the path in terms of Euclidean distance is also indicated at each iterations. Table I summarizes the number of iterations and time consumed by IB-RRT*, B-RRT* and RRT* to reach an optimal path in this problem. It should be noted that the RRT* algorithm is unable to fully sample the given configuration space and thus fails to converge to the optimal solution within the limit of 5 million iterations. Although both IB-RRT* and B-RRT* were successful in finding the optimal solution, B-RRT* took an extremely large number of iterations to converge in comparison with IB-RRT*. B-RRT* utilizes the partial greedy heuristic approach as discussed earlier (algorithm line), this significantly reduces its ability of convergence to optimal path solution. Figures 4 and 5 represent particularly challenging maze type of cluttered 2D test environment. The environment has been set up in such a way that the starting and goal regions, while placed close together, are separated by the maze. All algorithms were tested, figure 4(a) to figure 4(d) and 5(a) to figure 5(d) show the convergence from the initial path solution to the optimal path solution by IB-RRT* and RRT*, respectively. For determination of the optimal path, the IB-RRT* algorithm takes the least number of average iterations ($i_{\text{avg}}$=95192) as compared to B-RRT*($i_{\text{avg}}$=286721) and the extraordinarily large number of iterations taken up by RRT* ($i_{\text{avg}}$=1059268) as shown in Table 1.

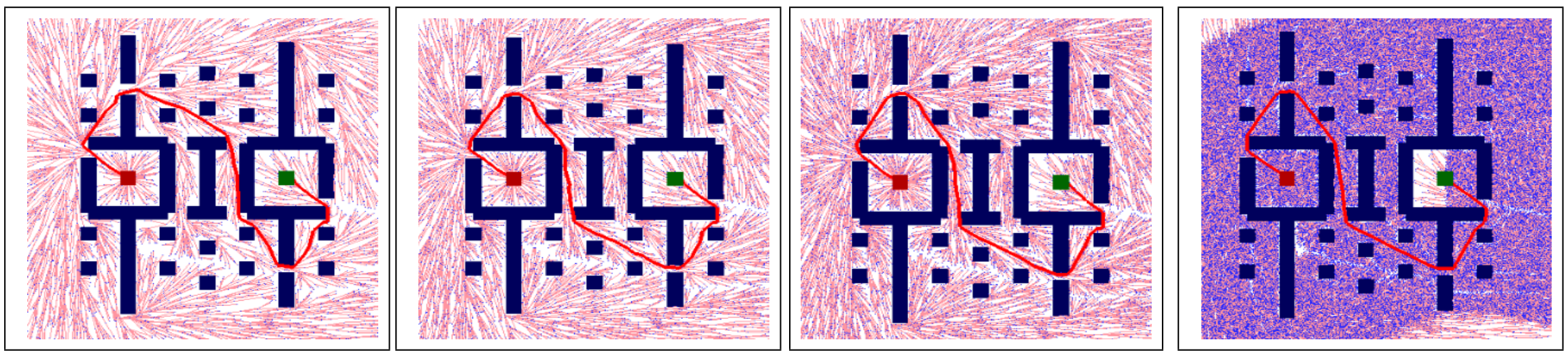

(a) i=31971,t=5.61s,C=72.4 (b) i=74119,t=12.81s,C=71.4 (c) i=317951,t= 55.6s,C=70.6 (d) i=1071861 ,t=184.2s,C=69.4

Figure 5: RRT* performance in 2-D environment (B)

Figure 8 shows the 3-D environment containing a multiple of barriers which separate the initial state and the goal region. IB-RRT* determines an optimal path most quickly ($i$=204321) as compared to B-RRT* ($i$=838692) and RRT* ($i$=1961825). Although all algorithms utilises uniform sampling heuristic, however, IB-RRT* maximizes the rewiring process per iteration due to intelligent sample insertion heuristic and hence quickly converges to the optimal path solution as compared to B-RRT* and RRT*. Figures 6 and 7 depict different scenarios in three-dimensional space. Their results are summarized in the Table 1. It can be seen that a similar trend is followed by the algorithms in all environments i.e., IB-RRT* rapidly converges to the optimal solution followed by B-RRT* and then RRT*. Moreover, in the maze problem depicted in Figure 6, RRT* was unable to sample the area close to the goal region even after an extremely large number of iterations while IB-RRT* was able to fully explore the space in a few thousand iterations.

Figure 9 summarizes the experimental test results performed in 10 different complex cluttered 2D and 3D environments for the comparison of IB-RRT*, B-RRT* and RRT*. The comparison is done in terms of: (a) iterations and time consumed to determine initial path as well as optimal path solution; (b) memory consumed in term of bytes for the determination of optimal path solution (c) convergence rate. From figure 9(a) to figure 9(e), it can be seen that IB-RRT* consumes lesser iterations, time and memory as compared to B-RRT* and RRT* for the determination of feasible path solution. Figure 9(f) provides another type of comparison using boxplot. In this the convergence rate of IB-RRT*, B-RRT* and RRT* are compared in these 10 different complex cluttered environments. Let the initial feasible path, denoted by $\sigma_{init}$, is computed in $t_{\text{init}}$ time while the optimal path solution, denoted as $\sigma*$, is computed in $t*$ time. Then the convergence rate is defined as $\frac{c(\sigma_{\text{init}}) - c(\sigma^*)}{t* - t_{\text{init}}}$. Since the process of convergence to the optimal path solution begins after finding initial feasible path solution, convergence rate is calculated after initial path computation. It is clear from the box plot that convergence rates of IB-RRT* are highest, followed by B-RRT* and RRT*. There also exists a sizable difference between the convergence rates of IB-RRT* and B-RRT*.

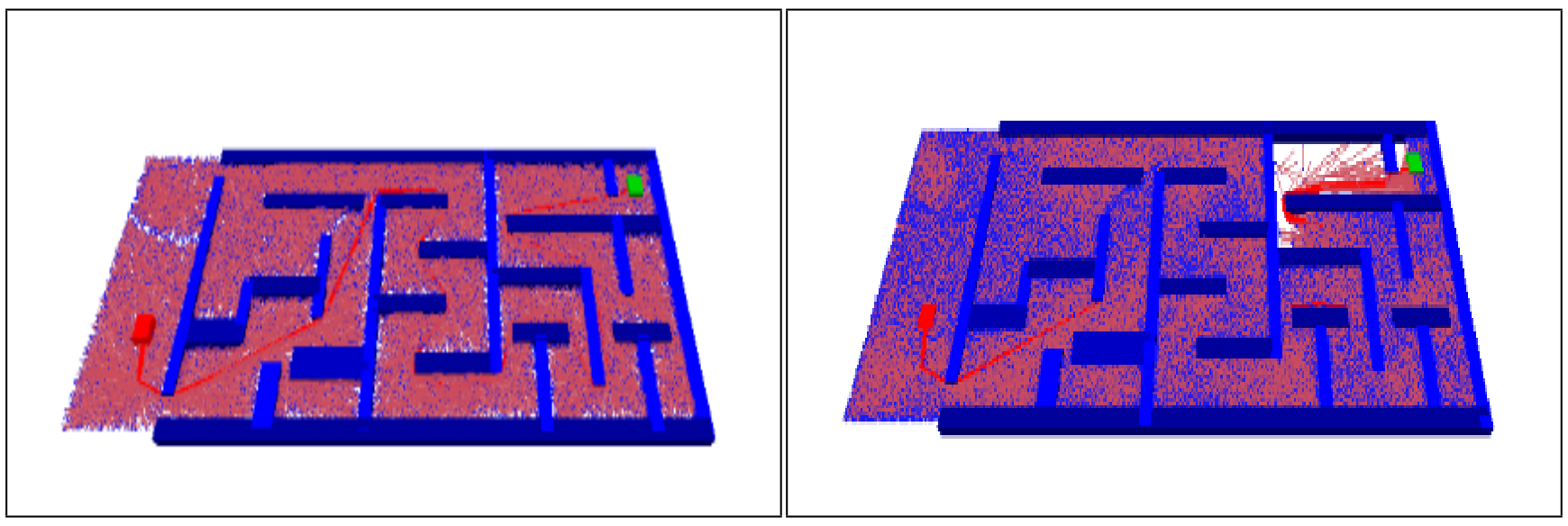

(a) IB-RRT*: i=162491,t=39.5s,C=299.2 (b) RRT*: i=2177186,t=376.4s,C=299.2

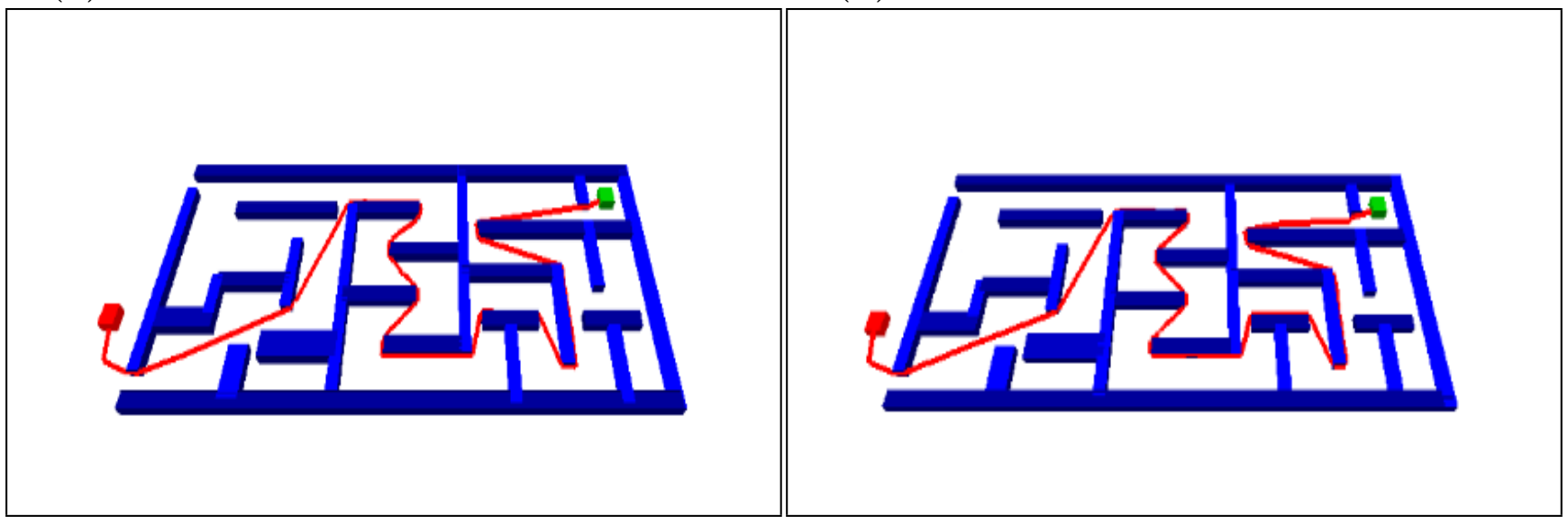

(c) IB-RRT*: Optimal path solution (d) RRT*: Optimal path solution

Figure 6: Performance of IB-RRT* and RRT* in Complex Maze Environment

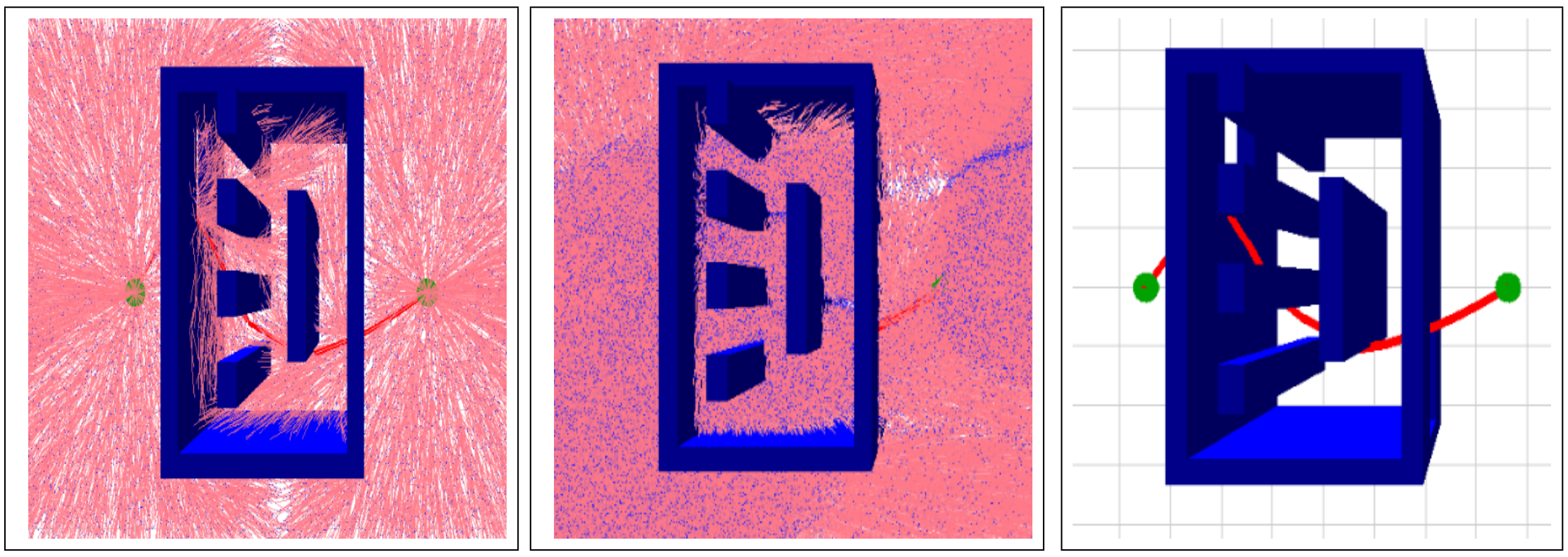

(a) IB-RRT*: i=79483,t=18.9s,C=69.8 (b) RRT*: i=1364129,t=235.9s,C=69.8 (c) Optimal path solution

Figure 7: Sequence of narrow passages

Figure 10 shows the running time ratio of a) IB-RRT* over BiRRT and b) IB-RRT* over RRT* after the execution of each iteration. The running time ratio of algorithm A (AL-A) over algorithm B (AL-B) is defined as the ratio of time consumed by AL-A over the time consumed by AL-B. It can be seen that as the number of iterations increases, the running time ratio reaches a constant value in both cases. Hence, large numbers of iterations imply that the random samples are fully and uniformly distributed in the obstacle-free space. However, before that time, computational complexity of IB-RRT* remains fairly lower than BiRRT and almost equal to RRT*. As a matter of fact, in this specific environment, the average amount of time taken by our proposed IB-RRT* algorithm to determine a viable path to the goal was seen to be barely four times that of BiRRT and 1.4 times that of RRT*.

# 8 Conclusions and Future work

This paper presents a detailed comparative analysis of performance of our proposed IB-RRT* algorithm with the existing algorithms RRT* and B-RRT*. It is proven both analytically and experimentally that our proposed algorithm i) has almost similar computational complexity as RRT* and BiRRT; ii) provides almost-sure convergence to the optimal path solution; iii) has the higher convergence rate meaning that it rapidly converges to the optimal solution as compared to both state of the art algorithms RRT* and B-RRT*; iv) consumes lesser memory to converge to the optimal solution, as it utilizes lesser iterations and each iteration consumes memory. This paper also presents path planning problems in which the original RRT* algorithm fails to reach the optimal path solution within reasonable limit of iterations. Experimental results supporting theoretical analysis are also presented in this paper. The proposed algorithm IB-RRT* allows rapid convergence to optimal solution without tuning of the sampling operation for optimal paths. Therefore, the proposed planner is of importance in the field of real time motion planning. Hence, we anticipate employing IB-RRT* for online motion planning of animated characters in complex 3-D environments.

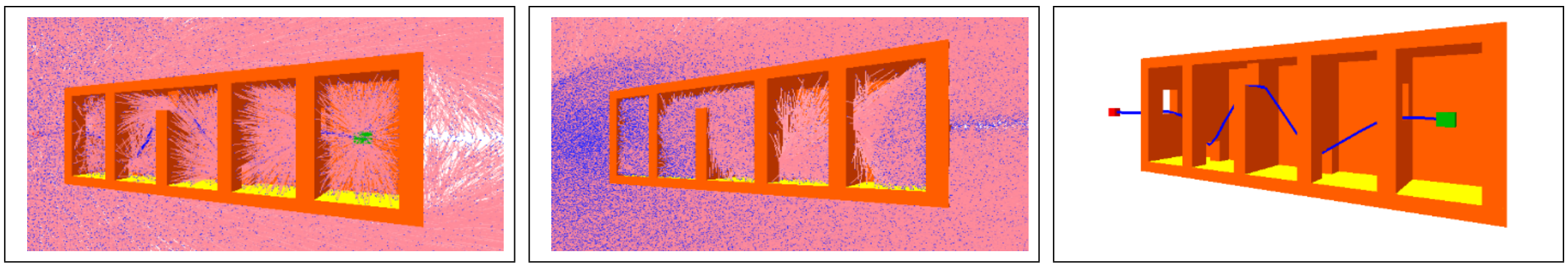

(a) IB-RRT*: i=186731,t=45.4s,C=81.9 (b) RRT*: i=2106391,t=364s,C=81.9 (c) Optimal path solution

Figure 8: Sequence of complex barriers.

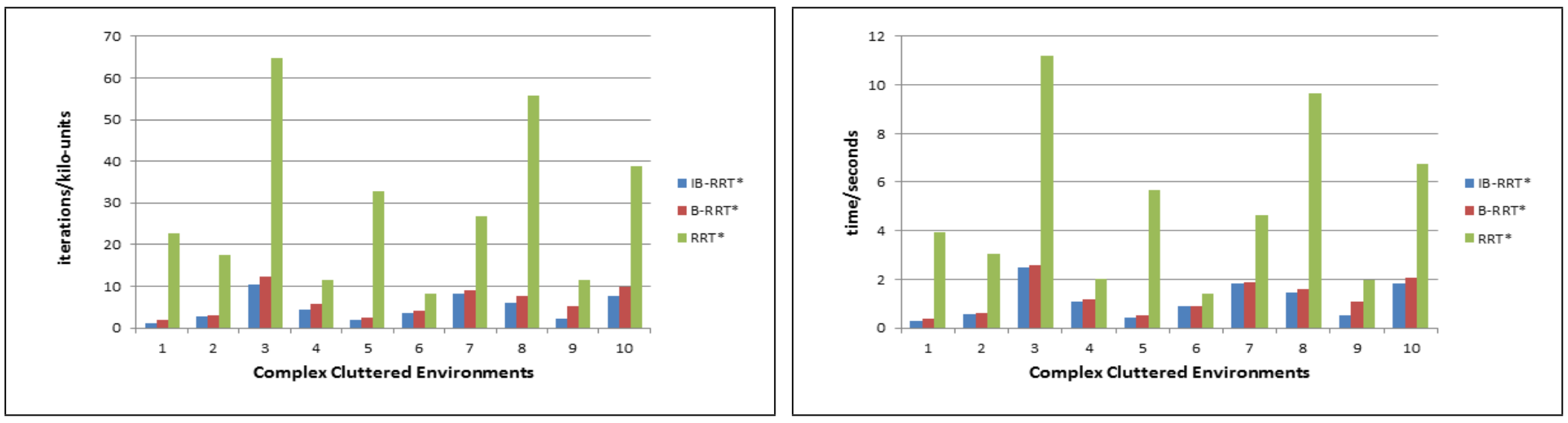

(a) Iterations used to find initial feasible path solution. (b) Time consumed to find initial feasible path solution.

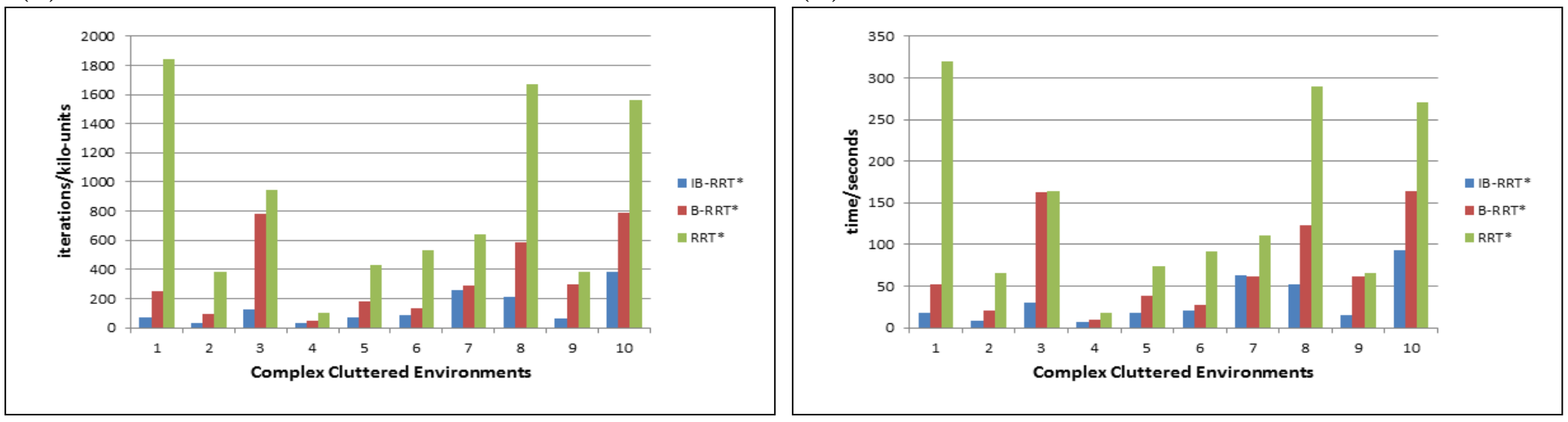

(c) Iterations used to find optimal path solution. (d) Time used to find optimal path solution.

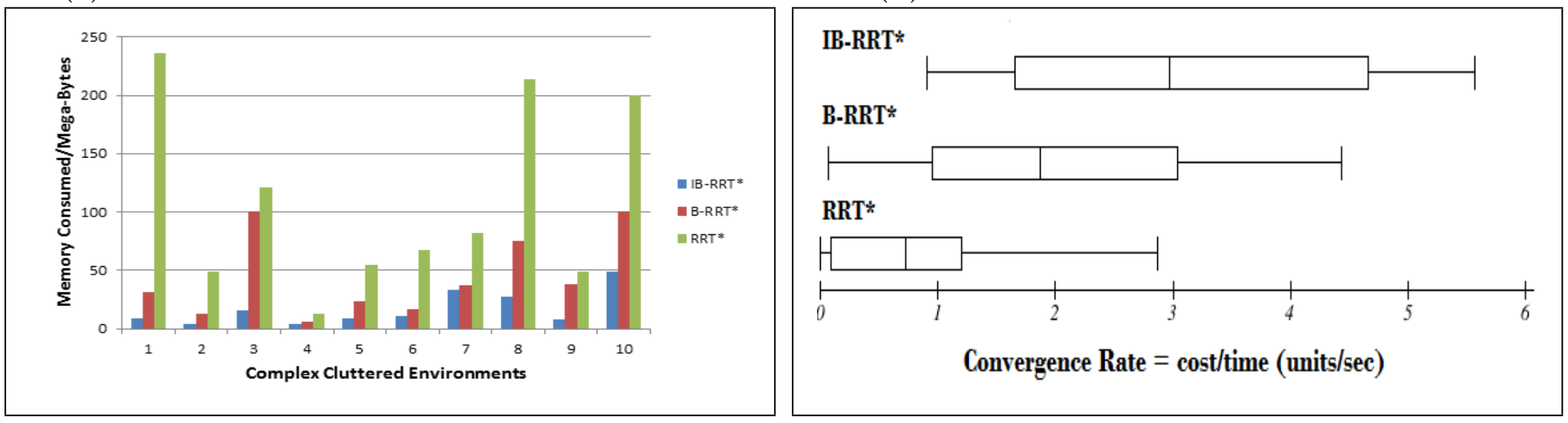

(e) Memory consumed to find optimal path solution. (f) Convergence rate comparison.

Figure 9: Comparison of IB-RRT*, B-RRT* and RRT* in 10 complex cluttered environments.

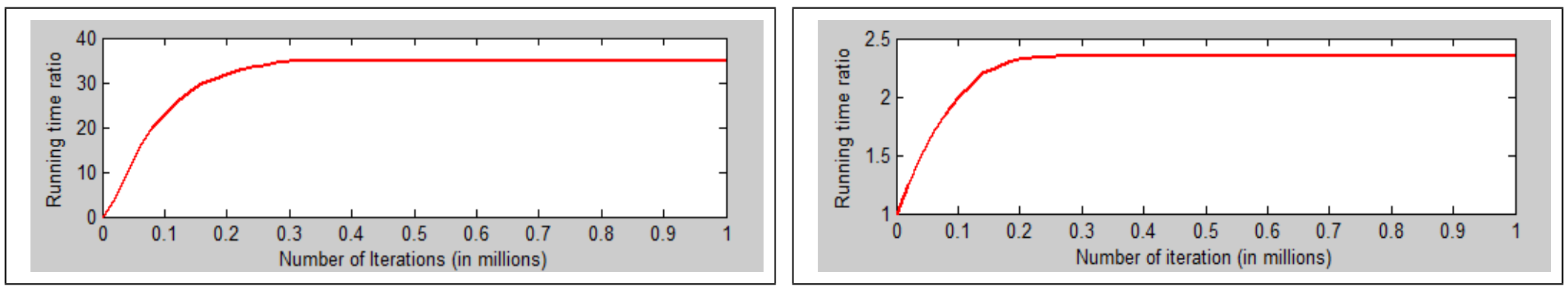

(a) (b)

Figure 10: Running time ratio of (a) IB-RRT* over BiRRT (b) IB-RRT* over RRT*.